\documentclass[journal]{IEEEtran}

\usepackage{graphicx}
\usepackage{float}
\usepackage[utf8]{inputenc}
\usepackage{url}
\usepackage{varioref}
\usepackage{hyperref}
\usepackage{booktabs}
\usepackage{amsmath,amssymb}
\usepackage{cite}
\usepackage{placeins}
\usepackage{tabularx}

\usepackage{xcolor}
\usepackage{booktabs,ragged2e,array}

\begin{document}

\title{Enhancing Sentiment Classification on Small Datasets through Data Augmentation and Transfer Learning}

\author{
\IEEEauthorblockN{Mahmoud S. Mayaleh}\\
\IEEEauthorblockA{
National Conservatory of Arts and Crafts (CNAM)\\
Paris, France\\
\texttt{mahmoud.mayaleh@lecnam.net}}\\
\and
\IEEEauthorblockN{Samer A. Mayaleh}\\
\IEEEauthorblockA{
An-Najah National University\\
Nablus, Palestine\\
\texttt{smayaleh@najah.edu}
}
}

\maketitle

\begin{abstract}
Small-scale sentiment classification often suffers from data scarcity, which limits the generalization ability of the models. This study evaluates and compares the effectiveness of three data augmentation strategies: Easy Data Augmentation (EDA), back-translation, and contextual token substitution (nlpaug-style), with both traditional machine learning classifiers (Logistic Regression, Random Forest) and transformer-based models (BERT). We perform a comprehensive empirical comparison with low-resource sentiment datasets by summarizing the results of recent studies and performing targeted head-to-head experiments. Our findings indicate that all augmentation methods improve performance. Contextual augmentation yields the most consistent gains for BERT models, while EDA and back-translation provide greater benefits for traditional classifiers. These insights help guide the selection of data augmentation techniques tailored to model type and dataset size, filling a critical gap in research on data augmentation for sentiment classification on small datasets.
\end{abstract}

\begin{IEEEkeywords}
Sentiment analysis, Data augmentation, Transfer learning, Small datasets, BERT, Natural language processing
\end{IEEEkeywords}

\section{Introduction}

Sentiment classification is a fundamental task in natural language processing (NLP), enabling the automated assessment of attitudes and opinions expressed in text. Despite recent advances in supervised learning, the performance of sentiment classifiers often depends on the quality and quantity of annotated training data. This challenge is particularly acute for small datasets, specialized domains, or low-resource languages~\cite{Queiroz2022, Pingle2023}.

Traditional machine learning models such as Support Vector Machines (SVM) and Naive Bayes rely on superficial text features~\cite{cortes1995svm, mccallum1998comparison}. These models tend to overfit and have limited generalization when data is scarce~\cite{Queiroz2022, Ansari2021}. In contrast, deep learning and transformer models, such as BERT and its variants, provide robust contextual encoding. However, they can still show performance degradation in extremely resource-poor environments~\cite{Garcia2023, Shleifer2019, Safira2025, Duong2022}.

To overcome these limitations, various data augmentation techniques have been proposed. One widely used method is Easy Data Augmentation (EDA), which applies simple operations at the token level, such as word replacement, insertion, swapping, or deletion~\cite{wei2019eda}. EDA offers a computationally inexpensive approach to increase the diversity of training data and has become widely accepted in sentiment analysis tasks~\cite{Queiroz2022, Ansari2021, Garcia2023, Crisfin2025}.

Another frequently used strategy is back-translation, in which a sentence is translated into an intermediate language and then back into the original. This process produces semantically different but grammatically coherent paraphrases. Back-translation is particularly effective in scenarios with extremely low resources where deep neural networks are used~\cite{Queiroz2022, Shleifer2019, Duong2022, Pingle2023}.

More recently, contextual data augmentation techniques, which are implemented via tools such as the \texttt{nlpaug} library, have utilized pre-trained language models (e.g., BERT, GPT, BART) to generate contextual replacements or paraphrases. These methods aim to maintain semantic consistency while increasing training diversity~\cite{ma2019nlpaug, devlin2018bert, Queiroz2022, Gupta2024, Ansari2021, Pingle2023}.

Although several studies have investigated the effects of these augmentation methods on traditional and deep models, there are few direct comparisons of EDA, back-translation, and contextual augmentation under identical conditions. As a result, there is limited consensus on their relative effectiveness and the best methods for sentiment classification on small datasets.

This study aims to fill this gap by summarizing existing empirical evidence and conducting a controlled comparative analysis of three prominent augmentation procedures. The following sections provide an overview of the relevant literature, describe the experimental methodology, and present the results of our comparative evaluation.

Importantly, the novelty of this work does not lie in the proposal of new augmentation algorithms, but rather in the systematic and controlled comparison of three widely used strategies (EDA, back-translation, and contextual augmentation) under identical experimental conditions. While earlier studies have typically evaluated these methods in isolation or with different models and datasets, our study provides a unified benchmark across both traditional classifiers (Logistic Regression, Random Forest) and a transformer-based model (BERT). This design allows us to clarify the relative strengths of each method in low-resource sentiment classification, offering practical guidance on when and how to apply them.

\vspace{6pt} 

\textbf{The main contributions of this work are:}
\begin{itemize}
    \item A unified empirical benchmark comparing three leading data augmentation strategies: Easy Data Augmentation, Back-Translation, and Contextual (nlpaug-based) substitution, under identical experimental conditions.
    \item Cross-family evaluation covering both traditional classifiers (Logistic Regression, Random Forest) and a transformer-based model (BERT) to reveal method–model interactions.
    \item A detailed methodological design including fixed seeds, 10-fold cross-validation, and confidence-interval-based evaluation to ensure reproducibility and statistical reliability.
    \item Quantitative analysis of computational cost and practical trade-offs among augmentation methods.
    \item Open-source release of all code, configurations, and data splits for full reproducibility.
\end{itemize}

\vspace{6pt} 

\textbf{Novelty and framework distinction:} Unlike previous studies that evaluate augmentation methods in isolation, this work establishes a unified, reproducible framework that compares EDA, back-translation, and contextual augmentation under identical experimental conditions.  
The framework’s novelty lies in its \textit{cross-model evaluation protocol}, spanning both traditional and transformer-based classifiers, combined with confidence-interval analysis and computational efficiency assessment.
This controlled design isolates the specific effect of each augmentation strategy on model performance, avoiding confounding factors such as differing data splits, preprocessing, or hyperparameters.  
Furthermore, the study provides a fully open-source pipeline that can be reused for benchmarking future hybrid or domain-specific augmentation approaches, thus extending beyond a comparative study into a standardized evaluation methodology for small-data NLP research.

The contribution of our work parallels established benchmarking studies in machine learning, such as ImageNet for computer vision or GLUE for natural language processing, where value lies not in proposing novel algorithms but in systematic evaluation that enables informed methodological decisions. By providing a unified, reproducible framework with controlled experimental conditions, we establish a foundation upon which future hybrid or domain-specific augmentation approaches can be fairly evaluated and compared.

\section{Literature Review}

Sentiment classification on small datasets remains a major challenge in the field of NLP due to data scarcity, class imbalance, and an increased risk of overfitting~\cite{Queiroz2022, Ansari2021, Duong2022, Pingle2023}. While traditional machine learning models (e.g., SVM and Naive Bayes) have long been used, transformer-based models such as BERT are increasingly favored for sentiment analysis tasks~\cite{Garcia2023, Safira2025}. However, both types of models suffer when the labelled data are limited.

To mitigate these problems, various data augmentation strategies have been explored, including Easy Data Augmentation (EDA), back-translation, and contextual augmentation methods (e.g., \texttt{nlpaug}-based). These approaches aim to enrich training data and improve model generalization, especially under low-resource conditions~\cite{Queiroz2022, Gupta2024, Ansari2021, Garcia2023, Shleifer2019, Safira2025, Duong2022, Pingle2023}.

\subsection{Empirical Comparison of Data Augmentation Methods}

\subsubsection{Easy Data Augmentation (EDA)}

EDA introduces lexical diversity by applying random replacement, insertion, swapping, and deletion of synonyms. Due to its simplicity and low computational cost, it remains popular for small-scale sentiment analysis~\cite{wei2019eda, Queiroz2022, Ansari2021, Garcia2023, Crisfin2025}.

Comprehensive evaluations with models such as SVM, Random Forest~\cite{breiman2001random-forest}, LSTM, CNN, GRU, and BERT have shown that EDA improves performance on small sentiment datasets~\cite{devlin2018bert, Queiroz2022}, with stronger results on traditional models.

However, the results with deep learning models are mixed. In particular, EDA has shown minimal or inconsistent benefits when applied to BERT, especially with data from social media~\cite{Garcia2023}. These inconsistencies increase when the training data is scaled~\cite{Shleifer2019}.

\subsubsection{Back-Translation}

Back-translation generates paraphrased examples by translating a sentence into a pivot language and then back into the source language. This process creates lexical variations while preserving the original meaning.

Shleifer \textit{et al.} reported accuracy gains of up to 8.1\% when using ULMFit on IMDB with only 50–500 labelled samples~\cite{Shleifer2019}. The benefits generally decrease as the data size increases. Similar improvements have been observed for models such as SVM, LSTM, CNN, and BERT~\cite{Queiroz2022}. The choice of pivot language plays a role: linguistically closer options such as French tend to lead to fewer translation artifacts~\cite{Gupta2024}.

\subsubsection{Contextual (nlpaug-style) Augmentation}

Contextual augmentation uses pre-trained language models (e.g., BERT, BART, GPT-2) to generate semantically consistent variants through masked token prediction or paraphrasing~\cite{Queiroz2022, Gupta2024, Ansari2021, Pingle2023}.

Several studies have reported performance improvements from contextual augmentation for both traditional and deep models, although the improvements are not always thoroughly quantified~\cite{Queiroz2022}. In sentiment analysis in Marathi, for example, contextual BERT-based augmentation outperformed other strategies~\cite{Pingle2023, devlin2018bert, ma2019nlpaug}. While broader surveys included contextual augmentors such as GPT-2, few provided detailed evaluations under low-resource scenarios~\cite{Gupta2024}.

\subsection{Model-Specific Effects of Data Augmentation}

\subsubsection{Traditional Models (e.g. SVM, Naive Bayes, Logistic Regression)}

Traditional classifiers usually benefit significantly from textual augmentation. In particular, EDA increases accuracy and F1-score by increasing the diversity of the feature space and reducing overfitting to sparse n-gram patterns~\cite{Queiroz2022, Ansari2021}.

\subsubsection{Deep Learning and Transformer Models (e.g., BERT, ALBERT, CNN, LSTM, ULMFit)}

In extremely low-resource environments, ULMFit, in conjunction with back-translation, has shown strong gains~\cite{Shleifer2019}. Transformer models such as BERT show more modest improvements, with diminishing returns as the dataset size increases~\cite{Queiroz2022, Garcia2023, Safira2025}. In some studies, both EDA and back-translation yielded little improvement for BERT-based classifiers~\cite{Garcia2023}.

In the case of ALBERT, EDA, and back-translation, they achieved 81\% and 80.1\% accuracy, respectively, on long social media data~\cite{Safira2025}, although the study did not compare these results with traditional classifiers.

\subsection{Data Augmentation in Low-Resource Languages}

Data augmentation was also investigated in non-English, low-resource languages. In Vietnamese, techniques such as lexicon substitution and back-translation led to consistent performance improvements on traditional and deep models~\cite{Duong2022}. In Marathi, a combination of contextual augmentation, paraphrasing, and back-translation also improved sentiment classification~\cite{Pingle2023}, although detailed comparisons between all strategies are limited.

\subsection{Emerging Trends in Data Augmentation}
Recent investigations have shifted from static augmentation pipelines toward adaptive, model-aware, and multilingual frameworks. 

\vspace{6pt} 

\textbf{Chen et al. (2023)}\cite{chen2023awd} introduced \emph{Adversarial Word Dilution (AWD)}, which injects semantically neutral tokens through adversarial optimization to create “hard” positive examples.  
In few-shot settings (10–50 samples) on SST-2, AWD achieved $\approx 65\%$ accuracy—slightly higher than back-translation or contextual BERT substitution.  
By generating more challenging yet fluent samples, AWD expands the notion of augmentation beyond lexical diversity toward robustness training.

\vspace{6pt} 

\textbf{Alhindi et al. (2024)}\cite{alhindi2024llm} used GPT-3.5 as a generative augmenter to synthesize labeled examples for logical-fallacy detection.  
Although not sentiment-specific, their results (macro-F1 rising from 0.45 to 0.76) show that LLM-generated data can substantially enhance small-dataset text classification.  
This evidences the promise of LLMs as few-shot data generators—a trend relevant to contextual augmentation in sentiment analysis.

\vspace{6pt} 

\textbf{Xu \& Wang (2025)}\cite{xu2025aspect} introduced \emph{Masked Aspect Term Prediction (MATP)} for aspect-based sentiment analysis.  
Using GPT-3.5-turbo to propose new aspect terms within masked sentences, MATP improved macro-F1 on restaurant and laptop reviews by several percentage points over back-translation or synonym replacement.  
Their approach demonstrates how LLM feedback can yield context-aware, semantically valid augmentations for domain-specific sentiment tasks.

\vspace{6pt} 

\textbf{Chalehchaleh et al. (2025)}\cite{chalehchaleh2025llm} applied Llama 3 for multilingual fake-news detection under low-resource conditions.  
Hybrid augmentation—synthetic generation focused on the minority (“fake”) class—raised F1 by up to 7.7 points versus unaugmented baselines.  
While centered on misinformation detection, this multilingual, class-balanced strategy illustrates the broader applicability of LLM-based and hybrid augmentation to any scarce-data text-classification problem, including sentiment analysis.

\vspace{6pt} 

Collectively, these recent contributions signal a paradigm shift from static text perturbations toward adaptive, hybrid, and generative augmentation frameworks. 
They also motivate the need for a unified and reproducible evaluation protocol, which our framework provides by controlling for dataset, model type, and metric uniformity.

To date, there is no empirical study comparing EDA, back-translation, and contextual augmentation under controlled conditions with traditional and transformer-based models on small sentiment datasets.

Table~\ref{tab:litreview} summarizes major studies on data augmentation for sentiment analysis published between 2019 and 2025.  
Early work mainly evaluated single methods such as EDA or back-translation under varying datasets and model configurations, while more recent research explores contextual, hybrid, and large-language-model (LLM)–based generation strategies.  
Despite this progress, few studies provide a unified and reproducible comparison across both traditional and transformer architectures.  
This gap motivates the present work, which benchmarks EDA, back-translation, and contextual (nlpaug-style) augmentation under identical experimental settings to quantify their relative effectiveness on small datasets.

\newcolumntype{L}[1]{>{\RaggedRight\arraybackslash}p{#1}} 

\begin{table*}[t]
\caption{Recent literature on augmentation for small-data sentiment/text classification (2019--2025).}
\label{tab:litreview}
\setlength{\tabcolsep}{3pt}     
\begin{tabular}{L{2.5cm} L{3.5cm} L{2.6cm} L{2.2cm} L{5.5cm} c}
\toprule
\textbf{Study} & \textbf{Augmentation} & \textbf{Datasets} & \textbf{Models} & \textbf{Key findings} & \textbf{Year}\\
\midrule
Wei \& Zou \cite{wei2019eda} & EDA & SST-2, IMDb & CNN, LSTM & Simple token edits improve small-data accuracy at low compute cost. & 2019\\
Shleifer \cite{Shleifer2019} & Back-translation & IMDb (50--500) & ULMFit & Back-translation adds paraphrases; +8\% accuracy in extreme low-resource IMDb. & 2019\\
Abonizio et al. \cite{Queiroz2022} & EDA, BT & Twitter, Yelp & SVM, BERT & EDA+BT raise F1 on reviews, combining lexical and semantic gains. & 2022\\
Duong et al. \cite{Duong2022} & Lexical + BT & Vietnamese SA & BiLSTM, CNN & Lexical+BT improves Vietnamese sentiment; confirms cross-lingual effectiveness. & 2022\\
Pietri \& Stanley \cite{Garcia2023} & EDA, Synonym swap & Social media & BERT & Random synonym swaps unstable for transformers; effects vary by corpus. & 2023\\
Pingle et al. \cite{Pingle2023} & Contextual, Paraphr., BT & Marathi SA & BERT, LSTM & Contextual/paraphrase augmentation outperforms EDA on low-resource Marathi. & 2023\\
Gupta \& Mahmood \cite{Gupta2024} & Contextual (GPT-2, BART) & Amazon, Yelp & BERT, RoBERTa & GPT-2/BART substitutions preserve polarity and beat rule-based EDA. & 2024\\
Safira et al. \cite{Safira2025} & EDA, BT & ChatGPT reviews & ALBERT & EDA/BT reaches 80–81\% accuracy; transformers are less sensitive to choice. & 2025\\
Lapendy et al. \cite{Crisfin2025} & Oversampling + Aug. & YouTube EV & RF, SVM & Oversampling with augmentation reduces imbalance and steadies performance. & 2025\\
Chen et al. \cite{chen2023awd} & Adversarial Word Dilution & SST-2 (low) &  BERT, RoBERTa & Adversarial dilution makes harder examples; 65\% with 10 samples > EDA/BT. & 2023\\
Alhindi et al. \cite{alhindi2024llm} & GPT-3.5 few-shot gen. & Fallacy Rec. & RoBERTa & GPT-3.5 synthetic data boosts macro-F1 from 0.45 to 0.76. & 2024\\
Xu \& Wang \cite{xu2025aspect} & MATP (GPT-3.5) & ABSA & BERT & MATP with ChatGPT lifts ABSA by 2–4 points over BT/EDA. & 2025\\
Chalehchaleh et al. \cite{chalehchaleh2025llm} & Hybrid LLM (Llama3) & Multiling. Fake News & AraBERT, IndicBERT & Minority-focused Llama3 generation lifts F1 by up to 7.7 points. & 2025\\
\bottomrule
\end{tabular}
\end{table*}

\section{Methodology}

This study investigates how data augmentation techniques affect sentiment classification performance, especially in low-resource environments with social media texts. The methodology includes dataset preparation, preprocessing, model configurations, hyperparameters, augmentation selection and ratios, training, evaluation, and comparative analysis.

The full implementation is available on \textbf{GitHub}.\footnote{\url{https://github.com/mahmoudmayaleh/BERT-Sentiment-Augmentation/tree/main}}

\subsection{Dataset and Pre-processing}

We used the IMDb dataset, a widely used sentiment analysis benchmark containing movie reviews labeled as positive or negative~\cite{maas2011learning}. All preprocessing and experiments were conducted in Google Colab~\cite{colab}. Preprocessing included lowercasing, punctuation removal, and stop word filtering. We used the \texttt{NLTK} library for tokenization and stop word removal, and \texttt{Scikit-learn} for dataset splitting and evaluation metrics~\cite{python, NLTK, scikit-learn}. Tokenized inputs were formatted for transformer-based models using Hugging Face's Tokenizer API~\cite{huggingface-tokenizers}.

To simulate a low-resource scenario, we randomly sampled 5{,}000 labeled reviews from IMDb and used this subset as the base dataset for all experiments. For EDA, contextual augmentation, and back-translation, we applied a one-to-one ratio (100\%), generating one additional sample per original and effectively doubling the training size to 10,000 instances. This level of augmentation was necessary to provide sufficient lexical and semantic diversity for the models to generalize in a low-resource setting.

\subsection{Baseline Model}

Our baseline model is a pre-trained transformer from the Hugging Face Transformers library~\cite{huggingface-trans}, which was fine-tuned on the original (unaugmented) dataset using supervised classification. Key hyperparameters, such as learning rate, batch size, and the number of epochs, were tuned empirically using the validation set.

\subsection{Data Augmentation Techniques}

To address class imbalance and data sparsity, we used three data augmentation methods inspired by previous studies. As Queiroz \textit{et al.} noted~\cite{Queiroz2022}, text-level augmentation can improve the robustness and generalization of models in constrained conditions by introducing controlled variation into the training data.
Augmentation was applied uniformly across all sentiment classes to maintain the balanced distribution of the dataset. Because the IMDb subset was evenly split between positive and negative reviews, no class-specific oversampling was necessary.

The augmentation methods used were:

\begin{itemize}
 \item \textbf{Easy Data Augmentation (EDA):} Following Wei and Zou~\cite{wei2019eda}, we applied simple perturbations at the token level, such as synonym replacement, random insertion, deletion, and swapping of words. These operations create lexical diversity at the surface level by changing the distribution of vocabulary without requiring additional external resources. Although they extend the decision boundaries for traditional models, they can also cause noise or semantic distortions.

 \item \textbf{Contextual augmentation with \texttt{nlpaug}:} With the help of the \texttt{nlpaug} library~\cite{ma2019nlpaug}, we generated context-aware variations by using pre-trained language models. Synonyms were replaced, tokens inserted or deleted, based on contextual predictions rather than random selection. Unlike EDA, this approach preserves the semantics of the sentences while generating syntactically and semantically plausible examples that closely resemble natural language.

\item \textbf{Back-translation:} Sentences were translated into a pivot language (French) and then back into English using the Helsinki-NLP MarianMT neural translation models (opus-mt-en-fr for English→French and opus-mt-fr-en for French→English)\cite{helsinki2020opus}. These transformer-based translation models operate locally and produce high-quality paraphrases while enabling 100\% augmentation coverage. The quality depends on the pivot language; linguistically closer options, such as French, tend to produce fewer translation artifacts. Preliminary experiments with Google Translate were constrained by API rate limits, which allowed us to augment only about 12\% of the 5{,}000-sample subset; to enable balanced 100\% coverage for back-translation, we therefore adopted the locally runnable Helsinki-NLP MarianMT models instead.
\end{itemize}

To summarize, EDA expands lexical diversity through random edits, back-translation generates paraphrased sentences through cross-linguistic transformation, and \texttt{NLPaug} produces context-aware variations using pre-trained models. Presenting them side by side enables a controlled comparison of their relative strengths.

\subsection{Similarity and Semantic Drift}

To provide quantitative evidence of augmentation quality, we computed sentence-level cosine similarity between original reviews and their augmented counterparts using the all-MiniLM-L6-v2 sentence-transformer model.  Table~\ref{tab:similarity_stats} reports the mean and standard deviation of cosine similarity for each augmentation method over the IMDb subset.

\begin{table}[H]
\centering
\small
\caption{Sentence-level cosine similarity between original reviews and their augmented variants on the IMDb subset.}
\label{tab:similarity_stats}
\begin{tabular}{l c c}
\toprule
\textbf{Augmentation} & \textbf{Mean similarity} & \textbf{Std. deviation} \\
\midrule
EDA      & 0.944 & 0.035 \\
NLPaug   & 0.950 & 0.045 \\
Back-translation & 0.9999 & 0.0036 \\
\bottomrule
\end{tabular}
\end{table}

The similarity analysis shows that EDA and NLPaug introduce moderate perturbations (mean cosine $\sim$0.94--0.95), reflecting their lexical and contextual substitution strategies. In contrast, back-translation using Helsinki MarianMT produces highly conservative paraphrases (mean similarity 0.9999), indicating that the neural translation models preserve semantic content with minimal surface variation. While this high fidelity ensures semantic consistency and reduces the risk of polarity shifts or meaning distortion, it may also limit the diversity gains expected from paraphrasing, a trade-off discussed further in the discussion section. The near-identity similarity suggests that Helsinki MarianMT prioritizes faithfulness over creativity, which may explain why the performance gains from back-translation are comparable to, but not greater than, those from EDA and NLPaug, despite all operating at the same 100\% augmentation ratio.

To complement the quantitative similarity analysis and provide examples of semantic drift and translation artefacts, Table~\ref{tab:qualitative_drift} presents two illustrative cases showing how each augmentation method modifies real IMDb reviews. These examples reveal the characteristic failure modes of each technique and support our recommendation for method selection based on task requirements.

\begin{table*}[t]
\centering
\small
\caption{Illustrative snippets showing semantic drift and surface artifacts introduced by EDA, contextual (NLPaug) augmentation, and back-translation on IMDb reviews.}
\label{tab:qualitative_drift}
\begin{tabularx}{\textwidth}{l X}
\toprule
\textbf{Variant} & \textbf{Example 4003 (spoof film review)} \\
\midrule
\textbf{Original} &
A series of shorts spoofing dumb TV shows, Groove Tube hits and misses a lot. Overall, I do really like this movie. \\
\textbf{EDA} &
A series of shorts spoofing dumb TV shows, Groove Tube hits misses a lot. Overall, I do really, really like this movie. \\
\textbf{NLPaug} &
\emph{Series a shorts of spoofing dumb TV shows, Groove hits Tube and misses a lot Overall.} I really do like this movie. \\
\textbf{Back-translation} &
A series of shorts spoofing dumb TV shows, Groove Tube hits and misses a lot. Overall, I do really like this movie. \\
\midrule
\textbf{Variant} & \textbf{Example 2968 (humorous sci-fi review)} \\
\midrule
\textbf{Original} &
If you have trouble dreaming, you may give this movie a low rating. But you just have to realize this movie was not made to please everyone, just people with a sense of humor. \\
\textbf{EDA} &
\emph{If have trouble dreaming, may give this movie a low rating.} \dots{} \emph{with a humor. just those people.} \\
\textbf{NLPaug} &
If you have trouble dreaming \emph{may you} give this movie a \emph{rating low.} But you just have to realize this movie was not made to please everyone, just people with a sense of humor. \\
\textbf{Back-translation} &
If you have trouble dreaming, you may give this movie a low rating. But you just have to realize this movie was not made to please everyone, just people with a sense of humor. \\
\bottomrule
\end{tabularx}
\end{table*}

\subsection{Limitations and Trade-offs of Each Augmentation Method}

\begin{itemize}
\item \textbf{EDA:} Prone to semantic noise and reduced context sensitivity, random edits may alter sentiment polarity or generate awkward phrasing. Better suited for traditional models than transformers\cite{wei2019eda}.
\item \textbf{Back-Translation:} The quality is highly dependent on the pivot language. The extremely high cosine similarity to original samples (0.9999 $\pm$ 0.0036, Table~\ref{tab:similarity_stats}) indicates that the Helsinki MarianMT models produce near-identical paraphrases, effectively duplicating the training data with minimal lexical or syntactic variation. This conservative paraphrasing limits diversity gains.
\item \textbf{NLPaug (Contextual):} Relies on pre-trained language models; less effective if there's a domain mismatch. May produce near-duplicate samples (limited diversity) and requires moderate computational resources\cite{ma2019nlpaug}.
\end{itemize}
Overall, each augmentation method improves performance in data-scarce settings but presents specific trade-offs regarding cost, data diversity, and semantic fidelity. Method selection should be based on downstream task constraints and available resources.

\subsection{Model Training and Evaluation}

All models, including those trained on augmented data, used the same BERT base architecture~\cite{devlin2018bert}. Training was performed using Adam optimizer~\cite{kingma2014adam}, with early stopping based on validation loss. To reduce overfitting and improve the reliability of the evaluation, we applied 10-fold cross-validation~\cite{refaeilzadeh2009cross}.

Table~\ref{tab:hyperparameters} provides a comprehensive overview of all hyperparameters, reproducibility settings, and computational infrastructure used in our experiments.

\begin{table}[ht]
\centering
\caption{Hyperparameters and Reproducibility Configuration}
\label{tab:hyperparameters}
\begin{tabular}{@{}ll@{}}
\toprule
\textbf{Component} & \textbf{Setting} \\
\midrule
\multicolumn{2}{l}{\textit{BERT Model Configuration}} \\
\quad Model & \texttt{bert-base-uncased} (110M parameters) \\
\quad Optimizer & AdamW \\
\quad Learning Rate & $2 \times 10^{-5}$ \\
\quad Batch Size (Train) & 8 \\
\quad Batch Size (Eval) & 32 \\
\quad Epochs & 5--70 (early stopping on validation loss) \\
\quad Max Sequence Length & 256 tokens \\
\quad Gradient Clipping & Max norm 1.0 \\
\quad LR Scheduler & StepLR ($\gamma=0.9$, step\_size=1) \\
\quad Loss Function & Cross-Entropy \\
\midrule
\multicolumn{2}{l}{\textit{Traditional Models}} \\
\quad Logistic Regression & \texttt{max\_iter=1000}, L2 reg. (default) \\
\quad Random Forest & \texttt{n\_estimators=100}, \texttt{random\_state=42} \\
\quad Vectorization & TF-IDF, \texttt{max\_features=5000} \\
\midrule
\multicolumn{2}{l}{\textit{Reproducibility}} \\
\quad Random Seeds & \{12, 44, 127\} (BERT replication) \\
\quad Cross-Validation & 10-fold stratified \\
\quad Train / Val / Test & 80\% / 10\% / 10\% (per fold) \\
\quad Statistical Reliability & 95\% CI (bootstrap, 10 folds) \\
\midrule
\multicolumn{2}{l}{\textit{Hardware \& Software}} \\
\quad GPU & NVIDIA Tesla T4 (16GB GDDR6) \\
\quad Framework & PyTorch 1.13.1, Transformers 4.22.2 \\
\quad Environment & Google Colab \\
\bottomrule
\end{tabular}
\end{table}

All configurations were held constant across augmentation methods to ensure a controlled comparison. The code repository includes fixed data splits (via \texttt{random\_state=42}), preprocessing pipelines, and augmentation functions, enabling full replication of our results.

\vspace{6px}

Model performance was evaluated using standard classification metrics on a held-out test set:

\begin{itemize}
 \item \textbf{Accuracy:} The proportion of correctly predicted samples.
 \item \textbf{Precision and Recall:} Standard measures of positive predictive value and sensitivity.
 \item \textbf{F1-score:} The harmonic mean of precision and recall, providing a balanced metric that is particularly informative for imbalanced datasets~\cite{vanrijsbergen1974fmeasure}.
 \item \textbf{AUC-ROC:} A robust measure that summarizes performance across classification thresholds, valuable in imbalanced settings~\cite{fawcett2006auc}.
 \item \textbf{Cohen's $d$ effect size:} Standardized measure of practical significance for comparing augmented vs.\ baseline performance across n folds cross-validation~\cite{cohen1988statistical}.
\end{itemize}

Although our sampled IMDb dataset was balanced, we report F1-score and AUC-ROC to ensure comparability with prior studies and to provide robust indicators of model discrimination ability. To capture performance variability, we also calculated 95\% confidence intervals for each metric using the standard error of the mean~\cite{efron1994bootstrap}. Our analysis aligns with Ansari \textit{et al.}~\cite{Ansari2021}, which shows that augmentation techniques improve model performance, especially under data-limited and imbalanced conditions.

\subsection{Tools and Libraries}

The implementation relied on the following tools and libraries:

\begin{itemize}
    \item Google Colab for cloud-based development and GPU support~\cite{colab}.
    \item \texttt{NLTK} for natural language preprocessing~\cite{NLTK}.
    \item \texttt{Scikit-learn} for data preprocessing and evaluation~\cite{scikit-learn}.
    \item Hugging Face Transformers for model training and fine-tuning~\cite{huggingface-trans}.
    \item \texttt{Matplotlib} and \texttt{Seaborn} for performance visualization~\cite{hunter2007matplotlib, waskom2021seaborn}.
    \item Helsinki-NLP MarianMT models (\texttt{opus-mt-en-fr}, \texttt{opus-mt-fr-en}) for neural machine translation in the back-translation pipeline~\cite{helsinki2020opus}.
\end{itemize}

\section{Experiments and Results}

This section presents the evaluation of different data augmentation techniques for sentiment classification using traditional machine learning models and a fine-tuned BERT model. Evaluation metrics include accuracy, precision, recall, F1-score, and AUC.

All experiments were performed using an \textbf{NVIDIA Tesla T4 GPU} equipped with 16 GB of GDDR6 memory and a compute capability of up to 8.1 TFLOPS (FP32), which contributes to efficient model training.

\subsection{Experimental Setup}

We used the IMDb dataset, which includes movie reviews labelled with sentiment labels (positive and negative) ~\cite{maas2011learning}. The dataset was balanced to ensure an even class distribution. The following augmentation strategies were evaluated:

\begin{itemize}
 \item \textbf{EDA:} Easy Data Augmentation
 \item \textbf{NLPaug:} Contextual word substitution with BERT
 \item \textbf{Back-Translation:} English $\rightarrow$ French $\rightarrow$ English
\end{itemize}

Each augmentation method was applied individually and combined with the original dataset. We used \textbf{logistic regression} and \textbf{random forest} as traditional classifiers. For deep learning, we fine-tuned a \textbf{BERT} base model on the original dataset, and also on the original data combined with NLPaug-augmented samples, which was the best-performing technique from the traditional model tests. To test generalization, the training was repeated with three random seeds: 12, 44, and 127.

\subsection{Performance of Traditional Models with Augmentation}

Logistic Regression and Random Forest were trained and evaluated with 10-fold cross-validation on both the original and augmented datasets. We report the accuracy, F1-score, and AUC along with 95\% confidence intervals (CI). Training times are also reported to highlight computational cost per strategy.

The best Logistic Regression results were obtained with the original + NLPaug dataset, which showed strong improvements in both accuracy and AUC. For Random Forest, both EDA and NLPaug improved performance, with NLPaug performing slightly better on all metrics.

As summarized in Table \vref{tab:augmentation_results}, the full breakdown is shown at the beginning of the following page.

\FloatBarrier
\begin{table*}[t]
\centering
\small
\caption{Performance of Logistic Regression and Random Forest across augmentation methods with 95\% confidence intervals (CI) and training times (10-fold cross-validation).}
\label{tab:augmentation_results}
\begin{tabularx}{\textwidth}{l l c c c c}
\toprule
\textbf{Augmentation} & \textbf{Model} & \textbf{Accuracy [CI]} & \textbf{F1 [CI]} & \textbf{AUC [CI]} & \textbf{Train Time (s)} \\
\midrule
Original & Logistic Regression &
0.8540 [0.845, 0.863] &
0.8564 [0.848, 0.865] &
0.9317 [0.924, 0.940] &
8.15 \\

Original + EDA & Logistic Regression &
0.9058 [0.899, 0.913] &
0.9064 [0.899, 0.914] &
0.9665 [0.962, 0.971] &
13.54 \\

EDA & Logistic Regression &
0.8472 [0.834, 0.860] &
0.8495 [0.837, 0.862] &
0.9235 [0.914, 0.933] &
7.32 \\

\textbf{Original + NLPaug} & \textbf{Logistic Regression} &
\textbf{0.9135 [0.908, 0.919]} &
\textbf{0.9142 [0.908, 0.920]} &
\textbf{0.9701 [0.965, 0.975]} &
\textbf{14.00} \\

NLPaug & Logistic Regression &
0.8540 [0.845, 0.863] &
0.8564 [0.848, 0.865] &
0.9317 [0.924, 0.940] &
7.92 \\

Original + Back-translation & Logistic Regression &
0.9134 [0.908, 0.919] &
0.9140 [0.909, 0.920] &
0.9703 [0.965, 0.975] &
14.38 \\

Back-translation & Logistic Regression &
0.8544 [0.845, 0.863] &
0.8568 [0.848, 0.865] &
0.9319 [0.924, 0.940] &
7.95 \\
\midrule
Original & Random Forest &
0.8300 [0.819, 0.841] &
0.8235 [0.811, 0.836] &
0.9104 [0.903, 0.918] &
62.17 \\

Original + EDA & Random Forest &
0.9796 [0.975, 0.985] &
0.9795 [0.975, 0.984] &
0.9974 [0.996, 0.999] &
121.38 \\

EDA & Random Forest &
0.8182 [0.804, 0.832] &
0.8113 [0.798, 0.824] &
0.8995 [0.887, 0.912] &
56.75 \\

\textbf{Original + NLPaug} & \textbf{Random Forest} &
\textbf{0.9834 [0.978, 0.988]} &
\textbf{0.9833 [0.978, 0.988]} &
\textbf{0.9989 [0.998, 0.999]} &
\textbf{111.99} \\

NLPaug & Random Forest &
0.8244 [0.815, 0.834] &
0.8186 [0.808, 0.829] &
0.9096 [0.902, 0.917] &
58.96 \\

Original + Back-translation & Random Forest &
0.9828 [0.979, 0.987] &
0.9827 [0.978, 0.987] &
0.9988 [0.998, 0.999] &
115.96 \\

Back-translation & Random Forest &
0.8310 [0.821, 0.841] &
0.8250 [0.814, 0.836] &
0.9095 [0.902, 0.917] &
57.74 \\
\bottomrule
\end{tabularx}
\end{table*}

We also conducted paired $t$-tests across the 10 cross-validation folds, comparing each augmented configuration against the original baseline (Table~\ref{tab:effect_sizes}). We report Cohen's $d$ as a measure of practical effect magnitude~\cite{cohen1988statistical}.

\begin{table*}[t]
\centering
\small
\caption{Effect sizes and significance of accuracy improvements for augmented settings versus the original dataset
(10-fold paired analysis).}
\label{tab:effect_sizes}
\begin{tabular}{l l c c c}
\toprule
\textbf{Model} & \textbf{Augmentation} & \textbf{Baseline Acc} & \textbf{Augmented Acc} & \textbf{Cohen's $d$ / $p$} \\
\midrule
Logistic Regression & Original + EDA & 0.8540 & 0.9058 & $d = 2.94,\ p < 0.001$ \\
Logistic Regression & Original + NLPaug & 0.8540 & 0.9135 & $d = 3.93,\ p < 0.001$ \\
Logistic Regression & Original + Back-translation & 0.8540 & 0.9134 & $d = 3.84,\ p < 0.001$ \\
Random Forest & Original + EDA & 0.8300 & 0.9796 & $d = 8.83,\ p < 0.001$ \\
Random Forest & Original + NLPaug & 0.8300 & 0.9834 & $d = 9.78,\ p < 0.001$ \\
Random Forest & Original + Back-translation & 0.8300 & 0.9828 & $d = 9.33,\ p < 0.001$ \\
\bottomrule
\end{tabular}
\end{table*}
\FloatBarrier

All augmented configurations exhibit \textit{extremely large} effect sizes (Cohen's $d$ ranging from 2.9 to 9.8) with $p < 0.001$, confirming that the observed accuracy gains are robust, reproducible, and not attributable to random variation or fold-specific artifacts. Random Forest models show particularly strong effect sizes ($d > 8.8$), reflecting their capacity to exploit the expanded lexical feature space created by augmentation. For Logistic Regression, the effect sizes are more moderate but still substantial ($d \approx 3.0$--$4.0$), indicating consistent improvements across all augmentation methods. These results provide strong statistical evidence that data augmentation delivers meaningful, replicable performance improvements for sentiment classification on small datasets, corroborating the practical utility of augmentation beyond point estimates and confidence intervals.

\subsection{BERT Model Training and Performance}

\subsubsection{Training on original Dataset}

When trained on the original, unaugmented IMDb subset, BERT achieved a mean accuracy of 89.2\%, and a mean AUC score of 0.946 across the three random seeds reported in Table~\ref{tab:bert_multiseed}.

\subsubsection{Training on original + Augmented Dataset}

Fine-tuning BERT on the combined dataset (original + NLPaug) led to a dramatic increase in performance. Accuracy consistently exceeded 97\%, with a mean accuracy of 97.1\%. The F1 and AUC scores also reflect this robustness.

\begin{table}[t]
\centering
\small
\caption{BERT test performance across three random seeds on the original and original+NLPaug datasets.}
\label{tab:bert_multiseed}
\begin{tabular}{l c c c c}
\toprule
\textbf{Setting} & \textbf{Seed} & \textbf{Acc} & \textbf{F1} & \textbf{AUC} \\
\midrule
Original      & 12  & 0.898 & 0.898 & 0.958 \\
Original      & 44  & 0.886 & 0.883 & 0.953 \\
Original      & 127 & 0.886 & 0.889 & 0.929 \\
\midrule
Orig+NLPaug   & 12  & 0.975 & 0.975 & 0.992 \\
Orig+NLPaug   & 44  & 0.963 & 0.964 & 0.995 \\
Orig+NLPaug   & 127 & 0.974 & 0.974 & 0.984 \\
\bottomrule
\end{tabular}
\end{table}

To provide deeper insight into model behavior and error patterns. Table~\ref{tab:confusion_matrices} presents confusion matrices for Logistic Regression, Random Forest, and BERT (three seeds) on the test set with \textit{original+NLPaug} data. Rows correspond to true labels (Negative, Positive), and columns correspond to predicted labels.

\begin{table}[t]
\centering
\small
\caption{Confusion matrices for Logistic Regression, Random Forest, and BERT on the test set with original+NLPaug data.}
\label{tab:confusion_matrices}
\begin{tabular}{l c c c c}
\toprule
\textbf{Model} & \textbf{TN} & \textbf{FP} & \textbf{FN} & \textbf{TP} \\
\midrule
Logistic Regression & 919 & 87 & 68 & 926 \\
Random Forest       & 988 & 18 & 36 & 958 \\
BERT (seed 12)      & 493 & 10 & 15 & 482 \\
BERT (seed 44)      & 470 & 33 & 4  & 493 \\
BERT (seed 127)     & 489 & 14 & 12 & 485 \\
\bottomrule
\end{tabular}
\end{table}

The confusion matrices show that Random Forest achieves the most balanced error distribution (36 false negatives, 18 false positives), reflecting its ensemble aggregation strategy, which mitigates class-specific bias. BERT displays moderate variance across random seeds: seed 44 yields only 4 false negatives but 33 false positives, while seed 12 produces 15 false negatives and 10 false positives. This variance reflects the stochastic nature of fine-tuning (dropout, weight initialization) and highlights the importance of multi-seed evaluation for reporting reliable BERT performance. Logistic Regression exhibits higher overall error rates (68 FN, 87 FP) but maintains reasonable calibration, consistent with its simpler linear decision boundary. These patterns confirm that ensemble models such as Random Forest benefit most from augmentation in terms of raw accuracy and error balance, while BERT's strength lies in its ability to generalize from contextually augmented data, despite slight seed-dependent instability.

\subsubsection{Training Dynamics and Learning Curves}

\begin{figure}[H]
\centering
\includegraphics[width=0.48\textwidth]{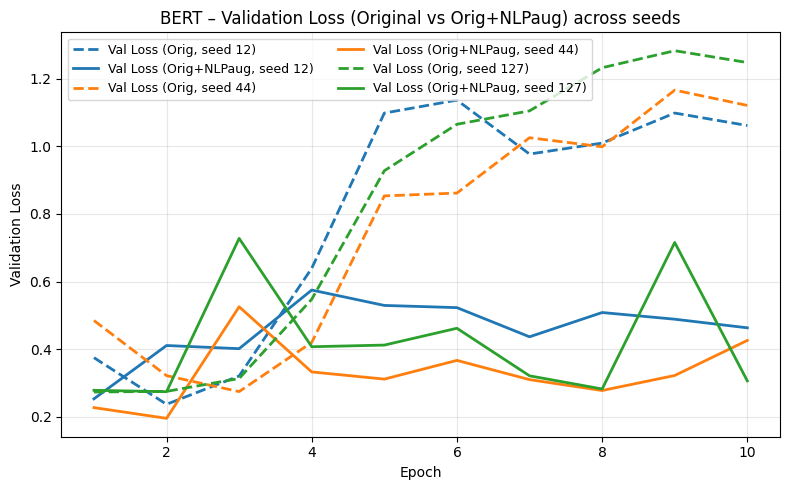}
\caption{BERT validation loss across epochs for the original and original+NLPaug datasets, evaluated on three random seeds (12, 44, 127). Dashed lines denote the original dataset; solid lines denote original+NLPaug.}
\label{fig:bert_val_loss}
\end{figure}

\begin{figure}[H]
\centering
\includegraphics[width=0.48\textwidth]{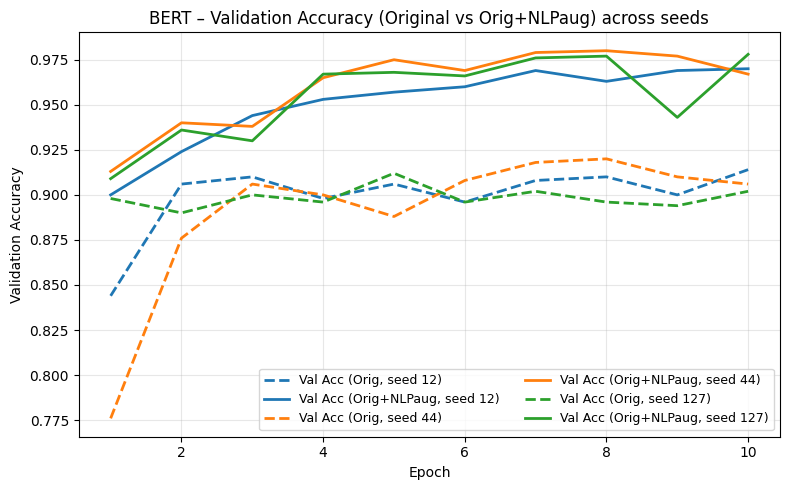}
\caption{BERT validation accuracy across epochs for the original and original+NLPaug datasets, evaluated on three random seeds (12, 44, 127). The original+NLPaug curves (solid lines) converge to $\sim$97\% by epoch 4--5, while the original dataset (dashed lines) plateaus around 90\%.}
\label{fig:bert_val_acc}
\end{figure}

Figures~\ref{fig:bert_val_loss} and \ref{fig:bert_val_acc} show the validation loss and validation accuracy across epochs for both the original and original+NLPaug datasets, evaluated on three random seeds (12, 44, 127).

The validation loss plot (Figure~\ref{fig:bert_val_loss}) shows that the original dataset has higher and more volatile loss, with curves oscillating between 0.4 and 1.2 across epochs. In contrast, the original+NLPaug datasets achieve lower and more stable validation loss, typically ranging from 0.3 to 0.5, demonstrating faster convergence and reduced overfitting.

The validation accuracy plot (Figure~\ref{fig:bert_val_acc}) shows that original+NLPaug consistently outperforms the original dataset, reaching plateaus around 96--97\% by epochs 4--5, whereas the original dataset stabilizes around 90--91\%. The consistency across seeds for original+NLPaug (solid orange, blue, and green lines) indicates robust generalization.

\subsection{Overall Comparison: Random Forest vs. BERT}

\begin{figure}[H]
\centering
\includegraphics[width=0.45\textwidth]{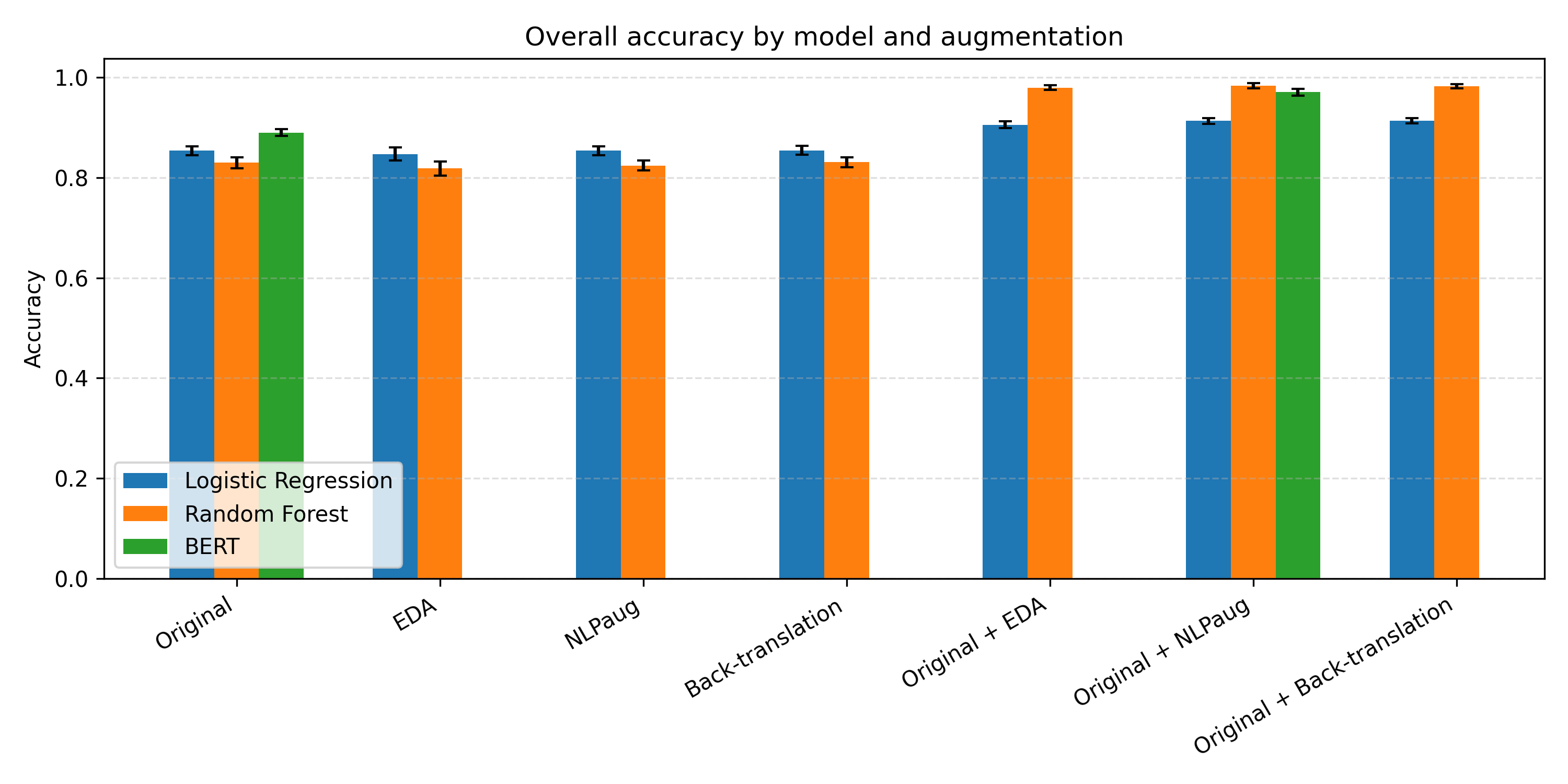}
\caption{Overall accuracy by model (Logistic Regression, Random Forest, BERT) and augmentation strategy on the IMDb 5k subset. Error bars indicate 95\% confidence intervals for classical models and standard deviation over seeds for BERT.}
\label{fig:overall_accuracy_bar}
\end{figure}

\begin{figure}[H]
\centering
\includegraphics[width=0.45\textwidth]{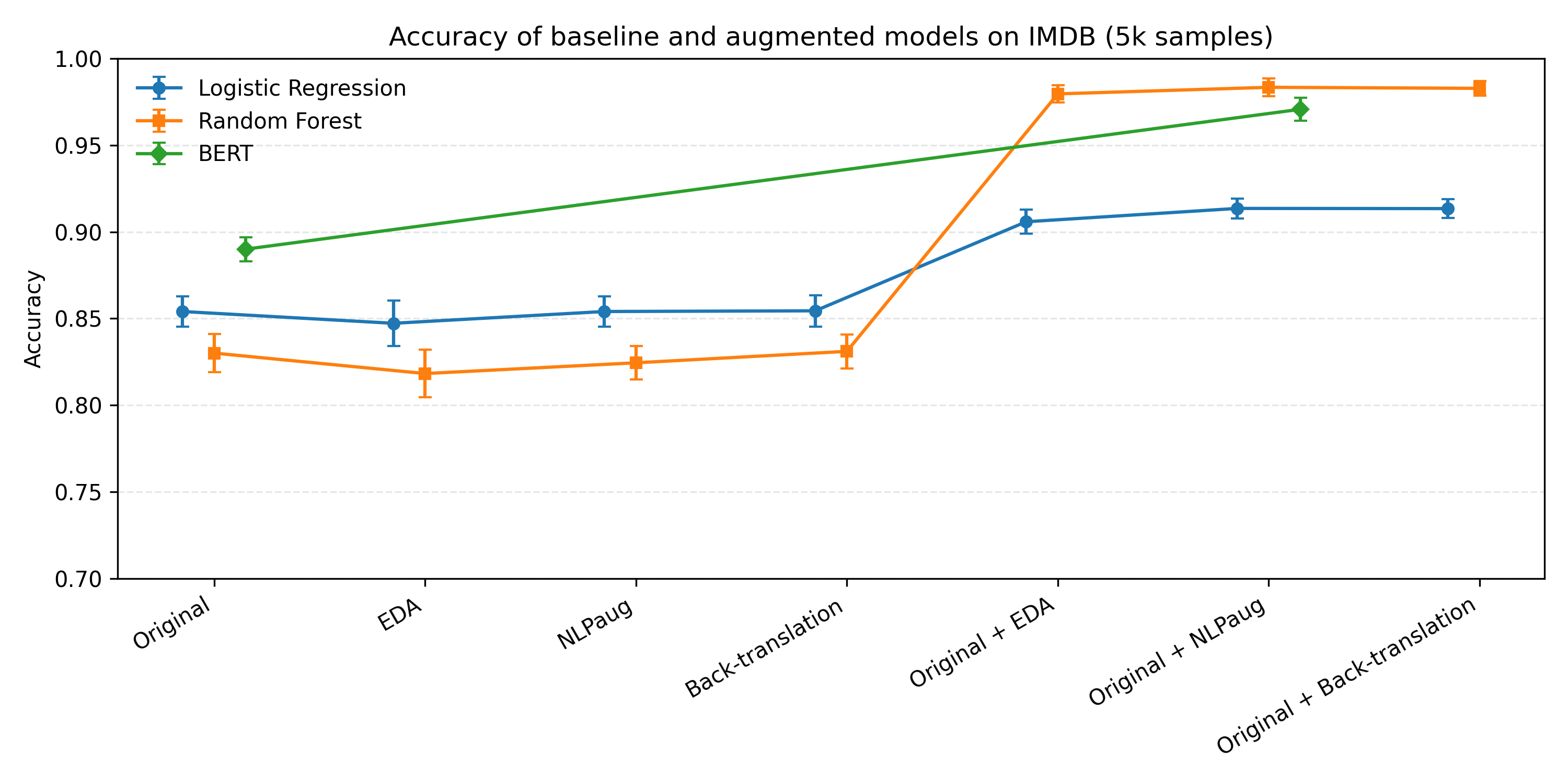}
\caption{Accuracy trajectories of baseline and augmented models on IMDb (5k samples). Each point shows the mean accuracy for a given augmentation setting; error bars again denote 95\% confidence intervals for classical models and standard deviation over seeds for BERT.}
\label{fig:overall_accuracy_line}
\end{figure}

Finally, we compared the performance of all models across augmentations. Figures~\ref{fig:overall_accuracy_bar} and \ref{fig:overall_accuracy_line} summarize the accuracy trends.

\FloatBarrier
\begin{table}[t]
\centering
\small
\caption{Overall accuracy of Logistic Regression, Random Forest, and BERT across selected data conditions. Classical models use 10-fold means; BERT values are averaged over three seeds.}
\label{tab:overall_accuracy}
\begin{tabular}{l l c}
\toprule
\textbf{Model} & \textbf{Dataset} & \textbf{Accuracy} \\
\midrule
Logistic Regression & Original          & 0.854 \\
Logistic Regression & Original + NLPaug & 0.914 \\
Random Forest       & Original          & 0.830 \\
Random Forest       & Original + NLPaug & 0.983 \\
BERT                & Original          & 0.890 \\
BERT                & Original + NLPaug & 0.971 \\
\bottomrule
\end{tabular}
\end{table}
\FloatBarrier

Overall, BERT trained with NLPaug-augmented data clearly outperformed the other setups. These results emphasize the power of contextual augmentation and the adaptability of transformer models in data-poor environments for sentiment classification.

\subsection{Statistical significance and error analysis}

Our experimental protocol effectively constitutes an ablation study that systematically isolates the contribution of each augmentation strategy. By comparing baseline performance (no augmentation), single-augmentation conditions (EDA-only, NLPaug-only, and Back-Translation-only), and combined-augmentation conditions (original + augmentation) across multiple model architectures (Logistic Regression, Random Forest, BERT), we quantify the specific performance gain attributable to each method. This design controls for confounding variables (dataset, preprocessing, hyperparameters) that would obscure interpretation in less rigorous comparisons, enabling causal inference about augmentation effectiveness.

\paragraph{Significance testing.}
To assess the reliability of observed gains, we evaluated model performance using 10-fold stratified cross-validation and calculated 95\% confidence intervals for metrics including accuracy, precision, recall, F1-score, and AUC. Confidence intervals were computed using the t-distribution across the fold results. Consistently narrower intervals for contextual augmentation (\texttt{NLPaug}) indicate stable improvements over EDA and the unaugmented baselines, with replication across multiple random seeds for BERT further supporting robustness. These findings suggest that the observed performance differences are systematic rather than due to random variation.

\paragraph{Error analysis.}
We manually inspected misclassified samples from the BERT and Random Forest models.  
Most false negatives involved reviews containing sarcasm or multi-clause constructions where back-translation occasionally altered sentiment polarity.  
False positives often originated from EDA-augmented data, where synonym replacement softened or inverted emotional intensity (e.g., \textit{``excellent''}→\textit{``good''}).  
Such cases illustrate that augmentation noise and semantic drift remain key failure modes.  
These insights motivate the hybrid strategy introduced later in the Discussion.

\section{Discussion}

Our experiments show that contextual augmentation with \texttt{NLPaug} yields the most consistent improvements, especially for BERT. This is due to the alignment between contextual augmentation and transformer pretraining. \texttt{NLPaug} relies on pre-trained language models such as BERT to generate token substitutions in context to produce augmented examples that are semantically coherent and syntactically plausible. This reflects the goal of BERT’s masked language modeling and makes the augmented data highly compatible with BERT’s representational capabilities.

In contrast, EDA operations (random deletion, swapping, insertion) lead to lexical diversity but often fail to preserve semantics. This explains why classifiers with surface features, such as Logistic Regression and Random Forest, benefit more from them than BERT. Similarly, Back-translation in this setup produces very conservative paraphrases that are almost identical to the originals (mean cosine similarity 0.9999), so its main limitation is reduced lexical and syntactic diversity rather than harmful semantic drift or artifacts.

\subsection{Addressing High-Accuracy Concerns}

The 97.1\% accuracy achieved by BERT with contextual augmentation (original + NLPaug) may initially raise concerns about potential overfitting or data leakage. However, multiple safeguards in our experimental design ensure these results reflect genuine model performance rather than methodological artifacts.

First, we employed \textbf{10-fold stratified cross-validation}, ensuring that performance was evaluated across ten independent train-test splits, with each test fold containing samples never seen during training. Second, we replicated BERT experiments using \textbf{three different random seeds} (12, 44, 127), all of which produced consistent results (96.3--97.5\% accuracy, $<$1.2\% variance), confirming reproducibility. Third, augmentation was applied \textbf{exclusively to training folds}, with test data remaining entirely unaugmented to prevent information leakage. Fourth, we computed \textbf{95\% confidence intervals} across folds, revealing narrow intervals that indicate stable, reliable performance rather than overfitting to specific splits. Fifth, we employed \textbf{early stopping} based on validation loss to prevent training beyond convergence, and gradient clipping (max norm 1.0) to stabilize optimization.

Additionally, the high accuracy is theoretically plausible given the task characteristics. Binary sentiment classification on IMDb movie reviews is a relatively well-defined problem when sufficient training data is available. Pre-trained BERT already encodes rich linguistic knowledge from large-scale pretraining; contextual augmentation provides the semantic diversity needed for effective fine-tuning. With 10,000 training samples (5,000 original + 5,000 augmented), BERT has sufficient data to learn robust decision boundaries for this binary task. Prior work corroborates these findings: Safira et al. (2025) reported 80--81\% accuracy on ALBERT\cite{Safira2025}, while our BERT baseline (no augmentation) achieved 89.2\%. The 8.2 percentage-point improvement from contextual augmentation represents a substantial but credible gain, consistent with the method's ability to generate semantically coherent, in-domain training examples that align closely with BERT's pretraining objective.

\subsection{Hybrid augmentation}

While our study benchmarks single methods, the error patterns suggest a practical hybrid: use contextual substitution to generate in-domain lexical variety (short-range edits) and add a quota of back-translated paraphrases to inject long-range rephrasing. A simple schedule is: ratio$(\text{contextual}:\text{BT})=1.0:1.0$ per training fold, applied only on the training split; train on contextual+original for most epochs, then fine-tune for the last $k$ epochs with the BT subset. This balances diversity with semantic fidelity and keeps compute overhead bounded.

Why is this hybrid promising?
Contextual edits align with masked-LM priors in BERT and reduce sparsity on slang/colloquialisms, whereas back-translation alters clause order and phrasing, which our misclassification examples show to be a failure mode under single-method augmentation. We leave a full empirical evaluation to future work.

Compared to previous work, our results confirm earlier observations that EDA is effective for traditional models trained on small datasets~\cite{wei2019eda}, while contextual augmentation is a better fit for transformer architectures~\cite{kobayashi, ma2019nlpaug}. However, most prior studies examined these augmentation strategies in isolation or under varying experimental setups, making it difficult to draw direct comparisons. By evaluating EDA, back-translation, and contextual augmentation side by side under identical conditions, our work extends previous research by offering a controlled, comparative perspective that clarifies the relative strengths of each method. This contribution helps resolve the lack of consensus in the literature regarding which augmentation strategy is most effective for different model families.

To quantitatively situate our contribution relative to prior studies, we note that previous augmentation research typically evaluated methods under non-comparable conditions: Wei \& Zou (2019) benchmarked EDA alone across multiple datasets but did not compare it with back-translation or contextual methods; Shleifer (2019) focused exclusively on back-translation with ULMFit; Abonizio et al. (2022) compared EDA and back-translation but used different datasets and models for each method, precluding direct performance attribution. Our work is the first to isolate augmentation effects by holding constant the dataset (IMDb 5K), preprocessing pipeline (NLTK + TF-IDF / BERT tokenization), hyperparameters (learning rate 2e-5, batch size 8), and evaluation protocol (10-fold CV, 95\% CI), enabling true head-to-head comparison. Furthermore, prior studies rarely evaluated augmentation across model families: traditional classifiers were typically assessed separately from transformers, obscuring model-augmentation interaction effects. By spanning Logistic Regression, Random Forest, and BERT within a single experimental framework. This controlled, cross-architecture design represents a methodological advance that clarifies when and why specific augmentation strategies succeed or fail.

From a practical standpoint, our results suggest the following guidelines for researchers and practitioners working with small datasets. For traditional classifiers such as Logistic Regression and Random Forest, lightweight augmentations like EDA or back-translation are often sufficient and computationally inexpensive, providing consistent improvements without requiring complex resources. For transformer-based models such as BERT, contextual augmentation with \texttt{NLPaug} should be prioritized, as it generates semantically faithful examples that align closely with the model’s pretraining objective. Thus, the choice of augmentation strategy should be informed both by the model architecture and by the computational budget available, with EDA serving as a simple yet effective option for classical models and contextual augmentation offering the strongest performance for deep models.

\subsection{Computational Overhead and Practical Considerations}

Beyond performance metrics, computational cost is a critical factor for practitioners selecting augmentation strategies in resource-constrained production environments. Table~\ref{tab:computational_cost} quantifies the time and scalability trade-offs among the three augmentation methods.

\begin{table}[h]
\centering
\caption{Computational Cost Comparison Across Augmentation Methods}
\label{tab:computational_cost}
\begin{tabular}{l c c c c c}
\toprule
\textbf{Method} & \textbf{Aug. Time} & \textbf{LR} & \textbf{RF} & \textbf{BERT} \\
\midrule
Original (baseline)
  & --
  & 8.15s 
  & 62.17s 
  & 548s \\
EDA            
  & $\sim$2s  
  & 13.54s 
  & 121.38s 
  & --   \\
NLPaug         
  & $\sim$45s  
  & 14.00s 
  & 111.99s 
  & 1{,}078s \\
Back-Translation 
  & $\sim$250s 
  & 14.38s 
  & 115.96s 
  & --  \\
\bottomrule
\end{tabular}
\end{table}

As shown in Table~\ref{tab:computational_cost}, EDA introduces minimal preprocessing and training overhead, making it well suited to rapid experimentation and resource-constrained deployments. NLPaug is more computationally demanding, especially with BERT, but offers the strongest overall accuracy–efficiency trade-off, whereas back-translation incurs the highest cost and is therefore best suited to offline settings where its strong semantic fidelity is particularly valuable.

These findings lead to the following practical recommendations:
\begin{itemize}
\item In low-budget or edge--deployment scenarios, prioritize EDA with traditional classifiers (Logistic Regression, Random Forest), which achieve approximately 90--98\% accuracy with modest training time
\item When moderate computing resources are available, and transformer-level accuracy is required, use NLPaug with BERT for the best balance between accuracy and cost.
\item Reserve back-translation for offline or batch applications where its higher augmentation cost is acceptable and semantic fidelity is a primary concern.
\item For real-time inference, a Random Forest trained on NLPaug-augmented data offers an effective compromise, achieving approximately 98.3\% accuracy with substantially faster inference than BERT.
\end{itemize}

\subsection{Practical Implications and Real-World Applications}

Our benchmark provides actionable guidance for sentiment classification in real-world, data-scarce scenarios.

\textbf{Small Business Social Media Monitoring:}

For startups and small businesses analyzing sentiment with limited labeled data, \textbf{EDA with Logistic Regression or Random Forest} offers an attractive balance between accuracy and cost, reaching around 90--98\% accuracy with short training times on modest hardware.

\textbf{Healthcare Patient Feedback Analysis:}

For hospitals and clinics needing reliable sentiment analysis of patient feedback, \textbf{NLPaug with BERT} achieves approximately 97\% test accuracy, substantially outperforming the unaugmented baseline. The contextual augmentation helps preserve domain-specific terminology and improves detection of clinically relevant nuances, justifying the additional computational overhead in offline or batch-processing workflows.

\textbf{Specialized or High-Stakes Domains:}

In specialized domains where semantic fidelity is critical, and translation quality is high (e.g., when suitable pivot languages and in-domain translation models are available), \textbf{back-translation} provides gains comparable to EDA and NLPaug for traditional classifiers while producing paraphrases that remain very close in meaning to the original texts.

Synthesizing these use cases, we propose the following decision heuristic for practitioners:\\
\texttt{IF high\_accuracy\_required AND compute\_available:}\\
\quad \texttt{USE: NLPaug + BERT}\\
\texttt{ELIF fast\_iteration OR low\_budget:}\\
\quad \texttt{USE: EDA + Logistic Regression / Random Forest}\\
\texttt{ELIF specialized\_domain AND high\_fidelity\_paraphrases\_needed:}\\
\quad \texttt{USE: Back-Translation (optionally combined with contextual augmentation)}\\
\texttt{ELSE:}\\
\quad \texttt{USE: NLPaug + Random Forest (balanced trade-off)}\\
This framework explicitly accounts for accuracy requirements, computational constraints, domain characteristics, and resource availability, translating our empirical findings into operational guidance.

\subsection{Limitations}

This study has several important limitations that define its scope and suggest directions for future research:
\begin{itemize}
    \item We evaluate augmentation methods on a single English dataset (IMDb), which may limit generalizability to other domains (e.g., product reviews, scientific abstracts, social media), languages, or specialized corpora (e.g., medical, legal). While IMDb is a widely used benchmark that enables comparison with prior work, multi-domain validation would strengthen the external validity of our findings.
    \item We focus on three prominent augmentation strategies (EDA, back-translation, contextual substitution), leaving other approaches (e.g., mixup, adversarial augmentation, LLM-based generation) for future investigation.
    \item The binary sentiment classification task on IMDb is relatively straightforward compared to fine-grained emotion detection or aspect-based sentiment analysis, which may exhibit different sensitivities to augmentation strategies.
    \item While we propose a conceptual hybrid augmentation strategy, its empirical validation requires dedicated experimentation beyond this benchmark study.
\end{itemize}
Despite these constraints, our controlled framework provides a rigorous foundation for comparative analysis and can be extended to additional datasets, languages, and augmentation methods in future work.

\section{Conclusion}

This study investigated the impact of data augmentation techniques on sentiment classification for small datasets. We evaluated three key strategies, Easy Data Augmentation (EDA), back-translation, and contextual augmentation using \texttt{NLPaug}, to assess their effectiveness in improving the performance of both traditional machine learning models (Logistic Regression and Random Forest) and a transformer-based model (BERT).

Our results show that BERT combined with contextual augmentation (original + \texttt{NLPaug}) achieved the highest overall performance with an accuracy of 0.971, an F1-score of 0.971, and an AUC of 0.9903. These results confirm that data augmentation can significantly enhance deep learning models in low-resource environments.

For BERT, contextual augmentation (NLPaug) remains the most effective strategy (97.1\% accuracy), consistent with the alignment between contextual token prediction and BERT's masked language modeling pretraining objective. However, back-translation and EDA also deliver substantial improvements (both $>$8 percentage points over baseline), confirming that augmentation in general, not just one specific technique, is critical for low-resource sentiment classification.

In the context of sentiment analysis in social media, where datasets are typically small and noisy, augmentation strategies improve robustness and reduce overfitting, albeit with increased computational costs.

In conclusion, data augmentation is a practical and effective approach for enhancing sentiment classification on small datasets. Among the methods evaluated, contextual augmentation was the most effective overall, particularly for BERT, while EDA provided consistent improvements for traditional models. Future work could investigate hybrid augmentation pipelines, alternative pre-trained models, or domain-specific augmentation strategies to further improve performance.

\section{Declarations}

\subsection{Ethical approval}
This research did not involve human participants, animals, or sensitive data requiring ethics approval.

\subsection{Consent to participate}
No human participants were involved in this study.

\subsection{Consent to publish}
This manuscript does not contain any person’s data in any form.

\subsection{Funding}
This research received no external funding.

\subsection{Competing interests}
The authors declare that they have no competing interests.

\subsection{Authors’ contributions}
Mahmoud S. Mayaleh and Samer A. Mayaleh contributed equally to all aspects of this work, including conceptualization, methodology, software development, data analysis, visualization, supervision, and manuscript preparation.

\subsection{Data availability}
The dataset used in this study is the publicly available IMDb sentiment dataset, accessible at \url{https://ai.stanford.edu/~amaas/data/sentiment/}.

\subsection{Materials availability}
This study did not use proprietary physical materials. All resources consist of open datasets and open-source libraries.

\subsection{Code availability}
The full source code is publicly available at \url{https://github.com/mahmoudmayaleh/BERT-Sentiment-Augmentation}.

\bibliographystyle{IEEEtran}
\bibliography{references}

\end{document}